\documentclass[conference]{IEEEtran}
\IEEEoverridecommandlockouts

\usepackage{cite}
\usepackage{amsmath,amssymb,amsfonts}
\usepackage[linesnumbered, ruled, vlined]{algorithm2e}

\SetCommentSty{mycommfont}
\usepackage{graphicx}
\usepackage{subfigure}
\usepackage{textcomp}
\usepackage{xcolor}
\usepackage{csquotes}
\let\oldnl\nl
\newcommand{\nonl}{\renewcommand{\nl}{\let\nl\oldnl}}
\DeclareMathAlphabet\mathbfcal{OMS}{cmsy}{b}{n}

\usepackage{multirow}

\begin{document}
	
	\title{Multi-task Optimization Based Co-training for Electricity Consumption Prediction
		\thanks{This work was supported in part by the Australian Research Council (ARC) under Grant No. LP180100114 and DP200102611.}
	}
	
	\author{\IEEEauthorblockN{Hui Song}
		\IEEEauthorblockA{\textit{School of Engineering} \\
			\textit{RMIT University}\\
			Melbourne, Australia \\
			hui.song@rmit.edu.au}
		\and
		\IEEEauthorblockN{A. K. Qin}
		\IEEEauthorblockA{\textit{Department of Computing Technologies} \\
			\textit{Swinburne University of Technology}\\
			Hawthorn, Australia \\
			kqin@swin.edu.au}
		\and
		\IEEEauthorblockN{Chenggang Yan}
		\IEEEauthorblockA{\textit{School of Automation} \\
			\textit{Hangzhou Dianzi University}\\
			Hangzhou, China \\
			cgyan@hdu.edu.au}}
	
	\maketitle
	
	\begin{abstract}
		
		Real-world electricity consumption prediction may involve different tasks, e.g., prediction for different time steps ahead or different geo-locations. These tasks are often solved independently without utilizing some common problem-solving knowledge that could be extracted and shared among these tasks to augment the performance of solving each task. In this work, we propose a multi-task optimization (MTO) based co-training (MTO-CT) framework, where the models for solving different tasks are co-trained via an MTO paradigm in which solving each task may benefit from the knowledge gained from when solving some other tasks to help its solving process. MTO-CT leverages long short-term memory (LSTM) based model as the predictor where the knowledge is represented via connection weights and biases. In MTO-CT, an inter-task knowledge transfer module is designed to transfer knowledge between different tasks, where the most helpful source tasks are selected by using the probability matching and stochastic universal selection, and evolutionary operations like mutation and crossover are performed for reusing the knowledge from selected source tasks in a target task. We use electricity consumption data from five states in Australia to design two sets of tasks at different scales: a) one-step ahead prediction for each state (five tasks) and b) 6-step, 12-step, 18-step, and 24-step ahead prediction for each state (20 tasks). The performance of MTO-CT is evaluated on solving each of these two sets of tasks in comparison to solving each task in the set independently without knowledge sharing under the same settings, which demonstrates the superiority of MTO-CT in terms of prediction accuracy.   
	\end{abstract}
	
	
	\section{Introduction}\label{introduction}
	
	Multi-task optimization (MTO)~\cite{song2019multitasking, swersky2013multi, chen2022multi}, a recently emerging research area in the field of optimization, mainly focuses on investigating how to solve multiple optimization problems at the same time so that the processes of solving relevant problems may help each other via knowledge transfer to boost the overall performance of solving all problems. MTO assumes some useful common knowledge exists for solving related tasks so that the helpful information acquired from addressing one task may be used to help solve another task if these two tasks have certain relatedness~\cite{liang2019hybrid}. Given its superior performance, MTO has been successfully applied to solve the benchmark optimization problems~\cite{gupta2017insights, qiao2022evolutionary, qiao2022dynamic} and real-world applications~\cite{zhang2018evolutionary, ye2019multitl, wu2020multi}. The research challenges arising from MTO include how to find the helpful source tasks for a target task and how the knowledge from selected source tasks can be extracted, transferred, and reused in a target task.      
	
	Evolutionary MTO (EMTO)~\cite{gupta2015multifactorial, liu2018surrogate} leverages evolutionary algorithms (EAs)~\cite{DEJONG2016} as the optimizer, aiming to unleash the potential of the implicit parallelism featured in EAs for solving MTO problems, where multiple optimization problems are addressed within a unified search space and knowledge is typically represented in the form of promising solutions and transferred via certain evolutionary operations such as crossover and mutation. The development of EMTO includes multifactorial evolutionary algorithm (MFEA)~\cite{gupta2015multifactorial} that is one of the most representative EMTO built on the genetic algorithm (GA), multitasking coevolutionary particle swarm optimization (MT-CPSO) that employs multiple swarms for solving multiple tasks~\cite{cheng2017coevolutionary}, an adaptive evolutionary multi-task optimization (AEMTO) framework that can adaptively choose the source tasks with probabilities for each target task working with differential evolution (DE)~\cite{xu2021evolutionary}, an evolutionary multitasking-based constrained multi-objective optimization (EMCMO) framework developed to solve constrained multi-objective optimization problems by incorporating GA~\cite{qiao2022evolutionary}, etc., from which different EAs are involved and their advantages are adopted to exchange knowledge among different tasks. 
	
	EMTO has been applied to address regression and classification problems~\cite{chandra2018evolutionary, chandra2018co}. A co-evolutionary multitasking learning (MTL) approach was proposed in~\cite{chandra2017dynamic} to solve a tropical cyclone wind-intensity prediction problem, where a multi-step ahead prediction problem is formulated as multiple one-step ahead prediction tasks with knowledge represented as a certain part of the neural network. A binary version of an existing multitasking multi-swarm optimization was proposed in~\cite{zhang2018evolutionary} to find the optimal feature subspace for each base learner in an ensemble classification model. In~\cite{shi2020evolutionary}, an evolutionary multitasking (EMT) ensemble learning model was proposed to solve the hyperspectral image classification problem by modeling feature selection (FS) as an MTO problem. An EMT-based FS method named PSO-EMT was proposed in~\cite{chen2021evolutionary} for solving the high-dimensional classification problem. PSO-EMT mainly focuses on converting a high-dimensional FS problem into several low-dimensional FS tasks and solving these tasks while enabling knowledge transfer between them.
	
	In this paper, we propose a multi-task optimization based co-training (MTO-CT) framework which trains multiple prediction models simultaneously, where an inter-task knowledge transfer module is designed to transfer and reuse knowledge (represented as model parameters) between these training tasks to facilitate solving them. The long short-term memory (LSTM)~\cite{hochreiter1997long} based model is employed as the predictor and optimized by a gradient descent (GD) based optimization method for all tasks. The predictor for each task has the same structure. In the inter-task knowledge transfer module, to decide which source tasks to be selected and the amount of knowledge within them to be transferred to help solve the target task, probability-based source task selection~\cite{xu2021evolutionary} is applied, where probability matching (PM)~\cite{thierens2005adaptive} is used to calculate the selection probabilities of all source tasks w.r.t. the current target task, and then stochastic universal selection (SUS)~\cite{baker1987reducing} is applied to select the most helpful ones from all sources tasks. Evolutionary operations are then applied to reuse the knowledge from the selected source tasks in the target task. Since this paper is to verify the superiority of MTO in addressing multiple tasks simultaneously, the proposed MTO-CT is compared with the single-task prediction (STP) model without knowledge transfer, i.e., solving each task in a standalone way, under the same settings.
	
	We use electricity consumption data from five states in Australia, i.e., VIC, NSW, SA, QLD, and TAS, to create two sets of tasks at different scales: a) one-step ahead prediction over five states (five tasks) and b) 6-step, 12-step, 18-step, and 24-step ahead prediction for each state (20 tasks), where electricity consumption data in different states share some common patterns. Also, in the multi-step ahead prediction problem, the next-step prediction depends on the knowledge of the previously predicted steps, which is an implicit form of common knowledge across different prediction tasks and makes it reasonable to regard prediction at different steps ahead as related tasks. In comparison to STP, the results on these two sets of tasks verify the superiority of MTO-CT.
	
	
	The rest of this paper is organized as follows. Section~\ref{problem_background} describes the problem formulated and the background knowledge. The proposed method and its implementation are presented in Section~\ref{method}. Section~\ref{result} reports and discusses experiments. Conclusions and some planned future work are given in Section~\ref{conclusion}.

	\section{Problem Definition and Background} \label{problem_background}
	This section will firstly introduce the problem defined. Then the background of LSTM is presented. 
	
	\subsection{Problem Definition}\label{problem_definition}
	Suppose there are $m$ time series $\textbf{X} = \{\textbf{x}_{1},..., \textbf{x}_{m}\}$, $\textbf{x}_{i} = \{x_{i, 1},..., x_{i, l_i}\}, \ i \in \{1, \dots, m\}$, where $l_i$ is the length of the $i^{th}$ time series. For any time series $i\in \{1, \dots, m\}$, there are $p$ different prediction purposes (e.g., different steps ahead prediction). An MTO-CT problem is defined as solving $n = mp$ prediction tasks at the same time. Given a predictor $h(\cdot)$, any prediction task $j \in \{1, \dots, n\}$ can be defined by $h_j(\widetilde{\textbf{x}}_{j};\mathcal{P}_j) \rightarrow \hat{\textbf{y}}_j$, where $\mathcal{P}_j$ denotes the parameter set of $h_j(\cdot)$ and $(\widetilde{\textbf{X}}, \textbf{Y}) = \{(\widetilde{\textbf{x}}_1, \textbf{y}_1), (\widetilde{\textbf{x}}_2, \textbf{y}_2),..., (\widetilde{\textbf{x}}_n, \textbf{y}_n)\}$ represents the training set for all $n$ task.

	Since the target task $j$ may benefit from addressing a source task $k \in \{1, \dots, n\}, k \neq j$ via knowledge transfer, knowledge from the source task (i.e., $\mathbfcal{P}_k \in \{\mathcal{P}_1, \mathcal{P}_2..., \mathcal{P}_n\}, \mathcal{P}_k \neq \mathbfcal{P}_j$) can be used to help boosting the prediction performance of the $j^{th}$ task. During the update of $\mathcal{P}_j$, knowledge from some selected source tasks based on certain probabilities according to their historical helpfulness is extracted, transferred, and reused to generate $\mathcal{P}_j^{new}$ for the $j^{th}$ target task to help improving its prediction performance.

	\subsection{Long Short-Term Memory}\label{background_lstm}
	
\begin{figure}[ht!]
		\centering
		\includegraphics[scale=0.6]{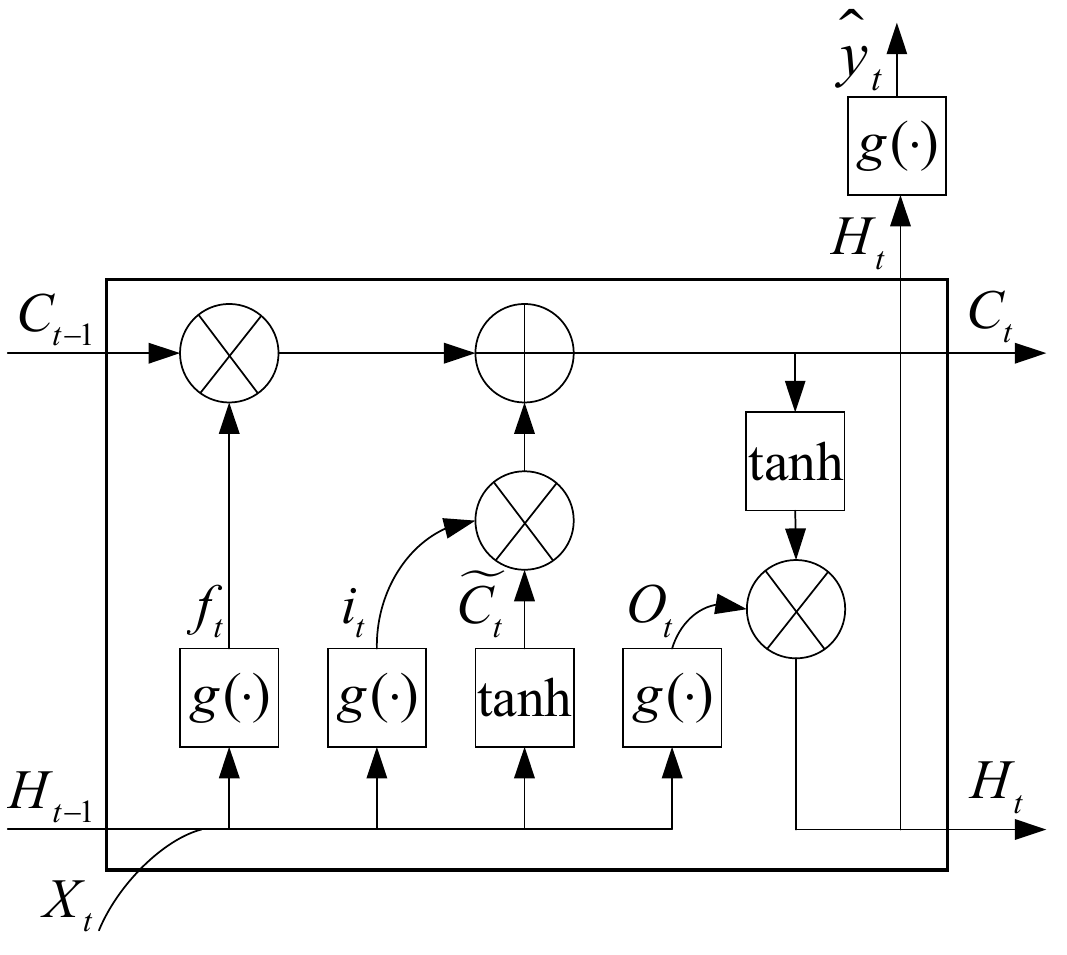}
		\caption{The structure of long short-term memory.}
		\label{lstm}
	\end{figure}

	Long short-term memory (LSTM), as a special kind of recurrent neural network (RNN), was proposed in  1997~\cite{hochreiter1997long} to overcome the shortcomings of recurrent backpropagation for learning to store information over extended time intervals. LSTM is explicitly designed to avoid the long-term dependency problem and remember information for long periods of time. Similar to the general RNNs, LSTM has a chain of repeating cells of an NN. The structure of an LSTM with one cell is illustrated as Fig.~\ref{lstm}, from which we can see there are a cell state ($C_{t-1}$) and three gates, i.e., forget gate ($f_t$), input gate ($i_t$), and output gate ($O_t$).     
	
	Given the inputs of current timestamp $X_t$, the hidden state and the cell state of the previous timestamp $H_{t-1}$ and $C_{t-1}$, three gates $f_t$, $i_t$, and $O_t$, and the next cell state $C_t$ can be obtained as follows:
	\begin{align}
		\centering
		\label{LSTM_gate}
		& f_t = g(W_f[H_{t-1}, X_t] + b_f]) \nonumber\\
		& i_t = g(W_i[H_{t-1}, X_t] + b_i])\nonumber\\
		& \tilde{C}_t = \tanh(W_c[H_{t-1}, X_t] + b_c])\nonumber\\
		& O_t = g(W_o[H_{t-1}, X_t] + b_o])\\
		& C_t = C_{t-1}f_t + i_t\tilde{C}_t\nonumber\\
		& H_t = O_t\tanh(C_t)\nonumber
	\end{align}
	
	In~(\ref{LSTM_gate}), $g(\cdot)$ in three gates is sigmoid function. With the current hidden state $H_{t}$, the prediction value $\hat{\text{y}}_t$ can be calculated according to:  
	\begin{align}
		\centering
		\label{LSTM_prediction}
		\hat{y}_t = g(W_yH_t + b_y)
	\end{align}
	
	The activation function $g(\cdot)$ in~(\ref{LSTM_prediction}) is sigmoid function in regression problems. The weights $W_f, W_i, W_c, W_o, W_y$ and biases $b_f, b_i, b_c, b_o, b_y$ over different cells are same. To obtain the optimal prediction result is to obtain the optimal weights and biases. 
	\begin{align}
		\centering
		\label{loss}
		\min \frac{1}{NT} \sum_{t=1}^{T}\sum_{s=1}^{N}L({y}_{t,s}, \hat{{y}}_{t,s})
	\end{align}
	
	The parameter set $\mathcal{P} = \{W_f, W_i, W_c, W_o, W_y, b_f, b_i, b_c, \\ b_o, b_y\}$ can be learned via any suitable optimization method using the loss function in~(\ref{loss}), where the real values are denoted as $\text{y}_t = \{y_{t, 1}, y_{t, 2},..., y_{t, N}\}$, $\hat{\text{y}}_t = \{\hat{y}_{t, 1}, \hat{y}_{t, 2},..., \hat{y}_{t, N}\}$ are the predicted values, $L(\cdot)$ is the evaluation method, $N$ represents the number of samples, and $T$ denotes the number of steps to be predicted. 
	
	\section{The Proposed Method}\label{method}
	
	We will first describe the proposed MTO-CT framework and then elaborate its inter-task knowledge transfer module responsible for selecting the most helpful source tasks in a probabilistic manner, adapting task selection probabilities, and reusing the knowledge from the selected source tasks to assist in the target task. We will also introduce an implementation of the MTO-CT framework.   
	
	\subsection{Framework}
	
\begin{figure}[ht!]
		\centering
		\subfigure[]
		{\centering\scalebox{0.9}
			{\includegraphics{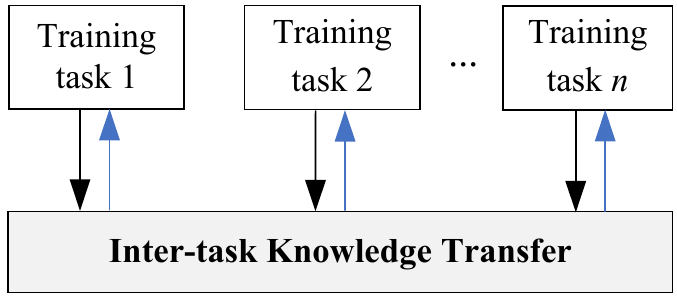}}\label{framework_a}}
		\subfigure[]
		{\centering\scalebox{0.9}
			{\includegraphics{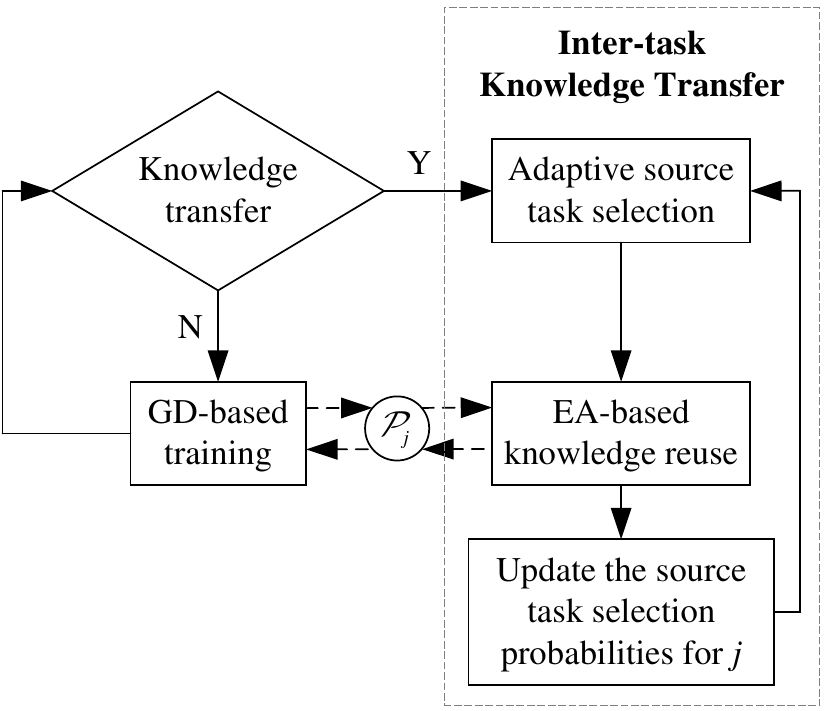}}\label{framework_b}}
		\caption{The illustration of MTO-CT framework: (a) the overall diagram and (b) the individual training process for each task $j, \ j \in \{1, \dots, n\}$.}
		\label{overall_framework}
	\end{figure}

	The proposed MTO-CT framework is illustrated in Fig.~\ref{overall_framework}, where Fig.~\ref{framework_a} shows the diagram of the co-training process with $n$ different tasks and Fig.~\ref{framework_b} describes the individual training process for each task $j, \ j \in \{1, \dots, n\}$. As shown in Fig.~\ref{framework_a}, all tasks are solved iteratively. In each iteration, each task is addressed independently with GD-based training before inter-task knowledge transfer is applied. After that, as illustrated in Fig.~\ref{framework_b}, if the knowledge transfer condition satisfies, e.g., for the $j^{th}$ task, the inter-task knowledge transfer will be applied. It first makes the adaptive source task selection, which consists of calculating the selection probabilities of $n-1$ source tasks according to their historical helpfulness in improving the performance of the $j^{th}$ task and selecting the candidates from these source tasks to extract knowledge. EA-based knowledge reusing mainly uses the operations in the EA to create the knowledge to be transferred and reused. The newly generated knowledge is utilized via updating $\mathcal{P}_j$. Finally, the effectiveness of the selected source tasks is quantified and used to update their selection probabilities for the next iteration. Notably, the MTO-CT framework can be treated as a special instance of the AEMTO framework~\cite{xu2021evolutionary}, where training instead of general optimization is incorporated. 
	
	\subsection{Inter-task Knowledge Transfer} 
	The inter-task knowledge transfer module consists of choosing the most helpful source tasks to help a target task based on their selection probabilities, transferring and reusing the extracted knowledge from the selected source tasks in the target task, and updating the selection probability of each source based on their helping performance. 
	
	\subsubsection{Source Task Selection}
	
	For any task $j, \ j \in \{1, \dots, n\}$, it has $n-1$ source tasks. In the inter-task knowledge transfer module in Fig.~\ref{framework_b}, the first step is to decide which source task(s) to be selected from the $n-1$ candidates. The source tasks that are more historically helpful may provide more useful knowledge. We calculate the probability of any source task $i$ according to its historical success rate in helping the target task $j$ iteration by iteration. We use $\textbf{q}_j = \{q_j^i|i = 1,...,n, i\neq j\}$ obtained from~(\ref{probability}) to denote the estimated helpfulness of each source task to the $j^{th}$ target task. With the obtained probabilities, the next step is to select $n_s$ source tasks from all $n-1$ candidates. We use SUS~\cite{baker1987reducing, xu2021evolutionary} with the source task probabilities $\textbf{q}_j$ to select $n_s$ source tasks for the target task $j, \ j \in \{1, \dots, n\}$. 
	
	
	
	\subsubsection{Knowledge Transfer and Reuse}
	
	After selecting $n_s$ source tasks via SUS, it is important to determine the amount of knowledge to be extracted from each of them and transferred to the target task $j, \ j \in \{1, \dots, n\}$, given that the source task with larger probability may provide more helpful knowledge to help solving the target task. We use the mutation operation used in differential evolution (DE)~\cite{qin2008differential} to generate a mutant $\mathcal{P}_{j}^{new}$ based on $n_s$ selected source tasks. In this work, we set $n_s = 3$ and adopt a popular DE mutation strategy~\enquote{DE/rand/1} to produce a mutant as follows: 
	\begin{align}
		\centering
		\mathcal{P}_{j}^{new} = \mathcal{P}_{j_1} + F \cdot (\mathcal{P}_{j_2} - \mathcal{P}_{j_3})
		\label{DE_rand_1}
	\end{align} 
	where $j_1, j_2, j_3 \in [1, n], j_1 \neq j_2 \neq j_3 \neq j$ denote three integers yielded via SUS. $F \in [0,1]$ is a positive real-valued control parameter for scaling the difference vector.
	
	To reuse the knowledge from the selected source tasks in the target task $j$, we apply the binomial crossover operation used in DE to the generated mutant and the target $\mathcal{P}_j$ to generate a new candidate solution as follows:
	\begin{align}
		\centering
		\mathcal{P}_{j}^{new, d} =
		\begin{cases}
			\mathcal{P}_{j}^{new, d}  & \text{if $rand_d[0, 1] \leq CR$} \\
			\mathcal{P}_{j}^d & \text{otherwise}
		\end{cases} 
		\label{crossover}           
	\end{align} 
	where $d \in \{1, \dots, D\}$ and $D$ denotes the number of elements in $\mathcal{P}_j, \ j \in \{1, \dots, n\}$ and $CR \in [0, 1] $ denotes the real-valued crossover rate. $\mathcal{P}_j^{new}$ and $\mathcal{P}_j$ will then compete for survival.
	
	\subsubsection{Source Task Selection Probability Update}
	
	The selection probability of each source task is initialized to a very small positive value. In each iteration, after reusing the knowledge from the $i^{th}$ source task in the $j^{th}$ target task, the corresponding helpfulness in the current iteration is measured via the reward $r_j^i$, which is then applied to update $q_j^i$ according to: 
	\begin{align}
		\centering
		q_j^i = \alpha q_j^i + (1 - \alpha)r_j^i
		\label{probability}
	\end{align}
	$r_j^i$ in this work is calculated by the successful rate of the $i^{th}$ task helping the $j^{th}$ task, i.e., $r_j^i = ns_j^i/(na_j^i + \epsilon)$, where $na_j^i$ and $ns_j^i$ denote the total number of times for the $i^{th}$ task selected to help the $j^{th}$ task over a certain period of time and the times that this help leads to the newly generated candidate solution to replace the target one. $\epsilon$ is a quite small positive value to avoid the issue of division by zero. 
	
	\begin{algorithm}[h!]
		\small
		\caption{Implementation of MTO-CT}
		\SetAlgoLined
		\label{MTO-CT_pseudocode}
		\KwIn{$(\widetilde{\textbf{X}}, \textbf{Y}) = \{(\widetilde{\textbf{x}}_1, \textbf{y}_1), (\widetilde{\textbf{x}}_2, \textbf{y}_2),..., (\widetilde{\textbf{x}}_n, \textbf{y}_n)\}$, $\textit{MaxIter}$, $\textit{CR} = 0.5$, $F = 0.2$, $n_s = 3$, $r_j^i=0$, $ns_j^i=0$, $na_j^i=0$, $q_j^i =0.005$, $j \in \{1, \dots, n\}$, $\ j \in \{1, \dots, n\}$, $i \neq j$, $\mathcal{T}=\{T_1, T_2,..., T_n\}$, $\alpha = 0.3, N, D, \#\textit{Iter}=0$}  
		\For{$j \rightarrow 1: n$}{Initialize the parameter set $\mathcal{P}_j$ from the standard normal distribution} 
		\While{$\#\textit{Iter} < \textit{MaxIter}$}{
			\For{$j \rightarrow 1: n$}{Evaluate the parameter set $\mathcal{P}_j$ on the $j^{th}$ task using~(\ref{RMSE}), denoted as $L_j$ \\ 
				\tcp{Inter-task knowledge transfer starts}
				\For{$i \rightarrow 1: n \ \& \ i\neq j$}{Calculate each source task selection probability according to~(\ref{probability}) to obtain the updated $q_j^i$}
				Perform SUS operation~\cite{baker1987reducing} to select $n_s$ source tasks, i.e., $j_1, j_2,..., j_{n_s}, k \in \{1, \dots, n_s\},  j_k \in \{1, \dots, n\}, j_k \neq j$\\ 
				Perform mutation operation according to~(\ref{DE_rand_1}) with the selected source tasks $j_1, j_2,..., j_{n_s}$ to obtain the $\mathcal{P}_j^{new}$ \\
				\For{$d \rightarrow 1: D$}{Perform crossover operation with~(\ref{crossover}) to update $\mathcal{P}_j^{new, d}$}
				Evaluate the newly generated $\mathcal{P}_j^{new}$ with~(\ref{RMSE}) to obtain the performance $L_j^{new}$ \\
				\If{$L_j^{new} < L_j$}{$\mathcal{P}_j^{new} \rightarrow \mathcal{P}_j$ \\
					$L_j^{new} \rightarrow L_j$\\
					$ns_j^{j_k}=ns_j^{j_k} + 1$}
				$na_j^{j_k}=na_j^{j_k} + 1, k \in \{1, \dots, n_s\}, \ j_k \in \{1, \dots, n\}$\\
				\tcp{Inter-task knowledge transfer ends}
				Update the parameter set $\mathcal{P}_j$ with Adam algorithm~\cite{kingma2014adam}} 
			$\#\textit{Iter} = \#\textit{Iter} + 1$}
		\KwOut{$\mathcal{P}_1^*, \mathcal{P}_2^*,..., \mathcal{P}_n^*$, $\hat{\textbf{y}}_1,\hat{\textbf{y}}_2,..., \hat{\textbf{y}}_n$}
	\end{algorithm}
	
	\subsection{Implementation}
	We implement the MTO-CT framework by using an LSTM-based prediction model for solving each of $n$ tasks. Given only a single time series is considered in this work, we adopt a less typical way to formulate the LSTM-based prediction task. Specifically, the input is defined as the time series values in a time window of $n_f$ consecutive timestamps and the output is defined as the time series values for $1,\dots,T_j$ steps ahead immediately following this window. Each LSTM cell has a single hidden layer and takes as inputs all $n_f$ time series values in the window as well as the hidden and cell states, where the first cell outputs the one-step ahead prediction, the second cell outputs the two-step ahead prediction and so on till the required $T_j$-step ahead prediction for the $j^{th}$ task is generated. As such, the total number of LSTM cells used is equivalent to $T_j$. The number of hidden neurons in a cell is set to $n_h$. This is different from a more typical way to formulating the LSTM-based prediction task in a~\enquote{many-to-many} manner, where each LSTM cell is fed in with only one time series value at a certain timestamp. 
	For each LSTM-based model, the parameters to be optimized (trained) include $\{W_f^j, W_i^j, W_c^j, W_o^j, W_y^j, b_f^j, b_i^j, b_c^j, b_o^j, b_y^j\}$ encoded via $\mathcal{P}_j$ of dimension size $D$.
	
	
	Each prediction task $j, \ j \in \{1, \dots, n\}$ is addressed by an adaptive moment estimation (Adam)~\cite{sun2019survey}, which is a first-order GD-based optimization method with the adaptive estimates of lower-order moments~\cite{kingma2014adam}. The inter-task knowledge transfer module is performed to explore if the information extracted from $n_s$ source tasks can further boost the prediction accuracy of the target task $j$. 
	
	We employ the root mean square error (RMSE) to define the loss function in~(\ref{loss}) for any task $j, j \in \{1, \dots, n\}$ as follows: 
	\begin{align}
		\centering
		\min \sqrt{\frac{1}{NT_j} \sum_{t = 1}^{T_j}\sum_{s = 1}^{N}({y}_j^{t, s}-\hat{{y}}_j^{t, s})^2}
		\label{RMSE}
	\end{align}
	where $T_j, j \in \{1, \dots, n\}$ denotes the time steps ahead to be predicted in the $j^{th}$ task, $\hat{\textbf{y}}_j$ denotes the predicted result w.r.t. the ground truth $\textbf{y}_j$, where $\hat{\textbf{y}}_j=\{\hat{\text{y}}_j^1, \hat{\text{y}}_j^2,..., \hat{\text{y}}_j^{T_j}\}, \hat{\text{y}}_j^t \in \Re^{1 \times N}$ and $\hat{\text{y}}_j^t = \{y_j^{t,1}, y_j^{t,2},..., y_j^{t,N}\}, t \in \{1, \dots, T_j\}$. 
	
	In each iteration, $\mathcal{P}_j$ is replaced by $\mathcal{P}_j^{new}$ via the inter-task knowledge transfer when $\mathcal{P}_j$ has worse performance than $\mathcal{P}_j^{new}$ in terms of~(\ref{RMSE}). This repeats until the maximum number of iterations ($\textit{MaxIter}$) is reached. The implementation of MTO-CT is detailed in Algorithm~\ref{MTO-CT_pseudocode}.

	\section{Results}\label{result}
	We will first present the data information and experimental settings. Then the prediction performance on these two different sets of tasks are presented and compared with STP to demonstrate the superiority of MTO-CT.

	\subsection{Data Description and Experimental Settings}
	
	The data is downloaded from Australian Energy Market Operator (AEMO)\footnote{http://www.nemweb.com.au/REPORTS/Archive/HistDemand/}. It includes electricity consumption data collected at 30-minute intervals from 01 November 2020 to 30 November 2021 for five states (VIC, NSW, SA, QLD, TAS) in Australia. We create two sets of tasks at different scales: (1) Set A: one-step ahead prediction across five states (five prediction tasks); (2) Set B: multi-step ahead (e.g., 6, 12, 18, 24) prediction for each of these five states (20 prediction tasks). For both sets, the time windows used as inputs are set as 24 (i.e., using the first 12 hours to predict the next step, next several steps, or the rest of the same day). For each of the tasks, there are 395 samples in total, where training and testing samples occupy 80\% (316) and 20\% (79), respectively. For each state, the data is normalized to $[0,1]$ using the min-max normalization. 
	
	The aim of this paper is to investigate if MTO can help improve the prediction accuracy when having multiple prediction tasks to be addressed simultaneously. We compare the results of MTO-CT with that of STP, which addresses every single task independently without inter-task knowledge transfer. The number of hidden neurons in LSTM is $n_h = 10$. In GD-based optimization method, i.e., Adam in this paper, learning rate is $l_r = 0.001$, $\beta_1 = 0.9$, $\beta_2 = 0.999$, $\epsilon = 1e-8$. To guarantee the comparison under the same number of evaluations, $\textit{MaxIter} = 20000$ in STP and $\textit{MaxIter} = 10000$ in MTO-CT considering the inter-task knowledge transfer operation in each iteration. All tasks are independently run ten times. 
	
	\begin{table}[!ht]
		\centering
		\caption{The task representations by the numbers.}
		\label{Task_number}
		\begin{tabular}{|cc|ccccc|}
			\hline
			\multicolumn{2}{|c|}{States}                              & VIC & NSW & SA & QLD & TAS \\ \hline
			\multicolumn{1}{|c|}{Set A}                  & one-step & 1   & 2   & 3  & 4   & 5   \\ \hline
			\multicolumn{1}{|c|}{\multirow{4}{*}{Set B}} & 6-step   & 1   & 2   & 3  & 4   & 5   \\
			\multicolumn{1}{|c|}{}                         & 12-step  & 6   & 7   & 8  & 9   & 10  \\
			\multicolumn{1}{|c|}{}                         & 18-step  & 11  & 12  & 13 & 14  & 15  \\
			\multicolumn{1}{|c|}{}                         & 24-step  & 16  & 17  & 18 & 19  & 20  \\ \hline
		\end{tabular}
	\end{table}

	\subsection{Results}
	
	The performance of MTO-CT is comprehensively studied by comparing it with STP over two sets of tasks, i.e., five and 20 tasks. The training and testing performance (RMSE) of these two sets of tasks over MTO-CT and STP is summarized and discussed, where the results is based on the normalized data. Wilcoxon signed-rank test at the 0.05 level is performed to estimate the significance of the difference between MTO-CT and STP. The better performance over the statistical test is labeled in bold. We use '+', '=', and '-' to indicate that the respective model has better, same, and worse performance than the other(s). To better understand what the task number represents in the following results, Table.~\ref{Task_number} gives the task representations with numbers.   
	
	\begin{table}[!ht]
		\centering
		\caption{Average RMSE over training and testing sets for one-step ahead prediction over five states ('+', '=', '-' : better, same, worse).}
		\label{result_5task}
		\begin{tabular}{|c|cc|cc|}
			\hline
			& \multicolumn{2}{c|}{Training RMSE} & \multicolumn{2}{c|}{Testing RMSE} \\ \hline
			Tasks & STP           & MTO-CT                    & STP           & MTO-CT                   \\ \hline
			1     & 0.05885       & \textbf{0.05747}       & 0.07012       & \textbf{0.06824}      \\ \hline
			2     & 0.0528        & \textbf{0.04966}       & 0.0592        & \textbf{0.05427}      \\ \hline
			3     & 0.05237       & \textbf{0.05141}       & 0.05283       & \textbf{0.0502}       \\ \hline
			4     & 0.06332       & \textbf{0.05859}       & 0.0588        & \textbf{0.05563}      \\ \hline
			5     & 0.05721       & \textbf{0.05457}       & 0.05918       & \textbf{0.05632}      \\ \hline
			+/=/- & 0/0/5  &    5/0/0       &  0/0/5    &     5/0/0           \\ \hline
		\end{tabular}
	\end{table}

	\subsection{Results of Set A}
	
	Table.~\ref{result_5task} reports the training and testing performance evaluated by mean RMSE over ten independent runs for MTO-CT and STP (without inter-task knowledge transfer), where the best mean RMSE for each task is labeled bold if it is significantly better with the statistical test. The result is based on one-step ahead prediction over VIC, NSW, SA, QLD, and TAS, respectively. By comparing the mean RMSE of MTO-CT and STP, it is obvious that MTO-CT outperforms both training and testing sets over all tasks from the labeled bold values and the total number of tasks it wins. This shows that the helpful knowledge reuse of the selected source tasks leads to significant improvement in the performance of the target task so that the accuracy of all tasks can be enhanced, which also demonstrates the effectiveness of inter-task knowledge transfer in MTO-CT.   
	
		\begin{figure}[h!]\centering
		\subfigure[STP]
		{\centering\scalebox{0.7}
			{\includegraphics{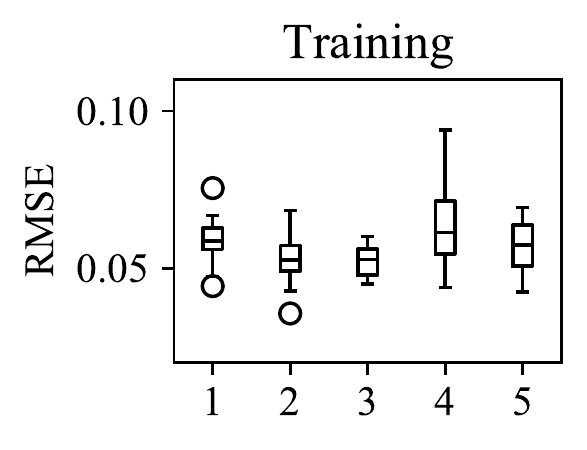}}\label{train_T5STL}}
		\subfigure[MTO-CT]
		{\centering\scalebox{0.7}
			{\includegraphics{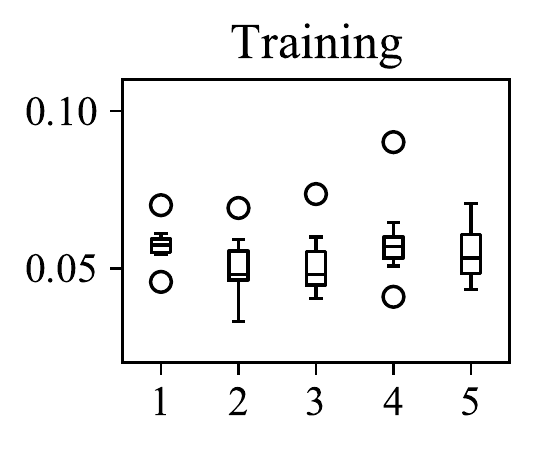}}\label{train_T5MTL}} \\
		\subfigure[STP]
		{\centering\scalebox{0.7}
			{\includegraphics{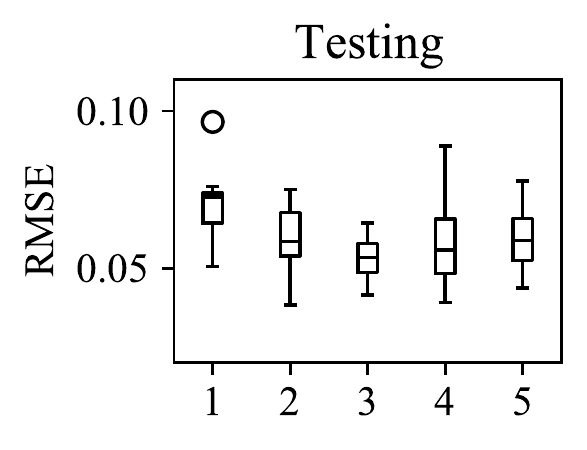}}\label{test_T5STL}}
		\subfigure[MTO-CT]
		{\centering\scalebox{0.7}
			{\includegraphics{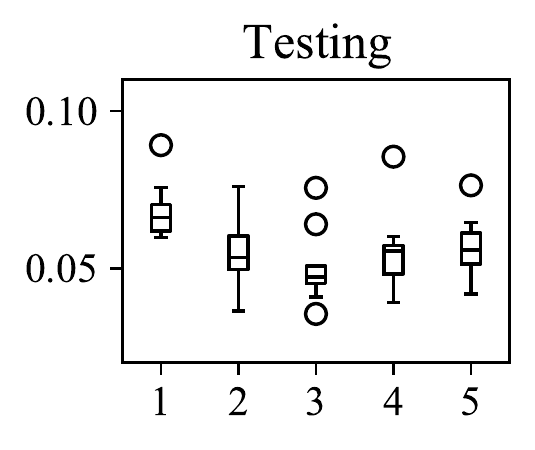}}\label{test_T5MTL}} 
		\caption{Box plots of training and testing RMSE on five tasks under STP and MTO-CT over ten independent runs.} \label{T5_distribution}
	\end{figure}
	
	\begin{table}[!ht]
		\centering
		\caption{Average RMSE over training and testing sets for five states across 6-step, 12-step, 18-step, and 24-step ahead prediction ('+', '=', '-' : better, same, worse).}
		\label{result_20task}
		\begin{tabular}{|c|cc|cc|}
			\hline
			& \multicolumn{2}{c|}{Training RMSE} & \multicolumn{2}{c|}{Testing RMSE} \\ \hline
			Tasks & STP                & MTO-CT               & STP               & MTO-CT               \\ \hline
			1     & 0.09486            & \textbf{0.09179}  & 0.10158           & \textbf{0.09783}  \\ \hline
			2     & 0.0903             & \textbf{0.0879}   & 0.09211           & \textbf{0.09006}  \\ \hline
			3     & \textbf{0.08879}   & \textbf{0.08655}  & 0.08504           & \textbf{0.08016}  \\ \hline
			4     & 0.09737            & \textbf{0.09588}  & 0.09111           & \textbf{0.08963}  \\ \hline
			5     & 0.07772            & \textbf{0.07625}  & 0.08044           & \textbf{0.07921}  \\ \hline
			6     & 0.12331            & \textbf{0.12401}  & \textbf{0.12088}  & 0.12204           \\ \hline
			7     & 0.11972            & \textbf{0.11756}  & 0.12252           & \textbf{0.11975}  \\ \hline
			8     & \textbf{0.11451}   & 0.11472           & 0.11011           & \textbf{0.10907}  \\ \hline
			9     & \textbf{0.13277}   & 0.13398           & \textbf{0.12709}  & 0.1276            \\ \hline
			10    & 0.09543            & \textbf{0.09385}  & 0.09304           & \textbf{0.09074}  \\ \hline
			11    & 0.12747            & \textbf{0.12587}  & 0.1283            & \textbf{0.12392}  \\ \hline
			12    & 0.12901            & \textbf{0.12548}  & 0.13309           & \textbf{0.12913}  \\ \hline
			13    & 0.12717            & \textbf{0.12657}  & 0.12319           & \textbf{0.12159}  \\ \hline
			14    & \textbf{0.13083}   & 0.13175           & 0.12711           & \textbf{0.12697}  \\ \hline
			15    & 0.10001            & \textbf{0.09781}  & 0.09744           & \textbf{0.09531}  \\ \hline
			16    & 0.1185             & \textbf{0.1177}   & 0.11515           & \textbf{0.11377}  \\ \hline
			17    & \textbf{0.11527}   & 0.1154            & 0.11907           & \textbf{0.119}    \\ \hline
			18    & \textbf{0.11751}   & 0.12034           & \textbf{0.11562}  & 0.11859           \\ \hline
			19    & 0.13719            & \textbf{0.13468}  & 0.13418           & \textbf{0.13309}  \\ \hline
			20    & 0.10684            & \textbf{0.10257}  & 0.10331           & \textbf{0.10123}  \\ \hline
			+/=/-   &5/1/14   & 14/1/5   &  3/0/17   & 17/0/3            \\ \hline
		\end{tabular}
	\end{table}
	
	\begin{figure}[h!]\centering
		\subfigure[STP]
		{\centering\scalebox{0.7}
			{\includegraphics{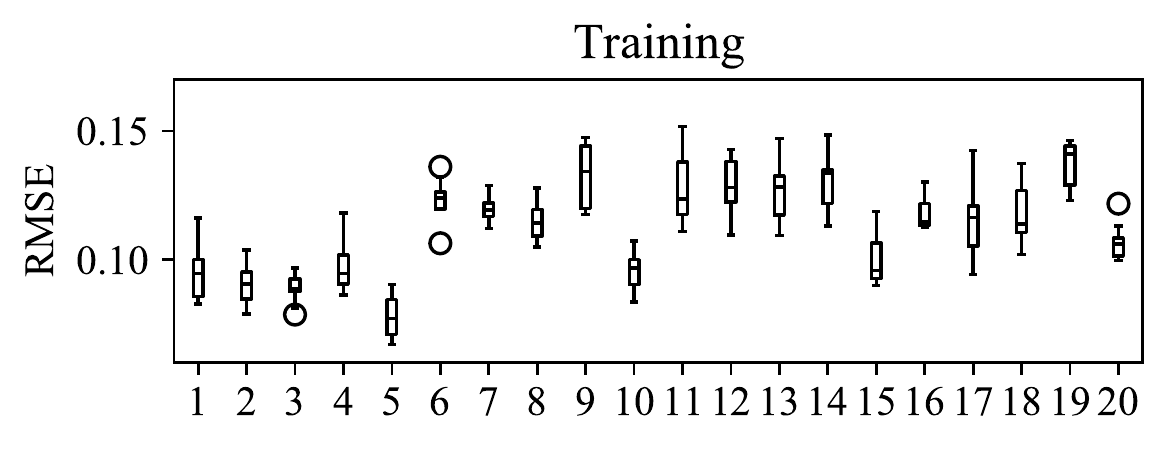}}\label{train_T20STL}}
		\subfigure[MTO-CT]
		{\centering\scalebox{0.7}
			{\includegraphics{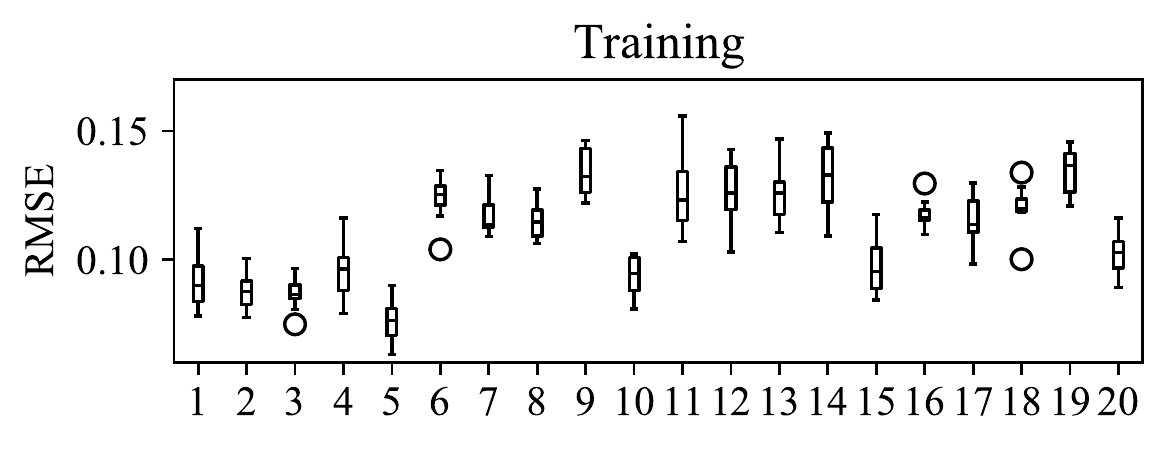}}\label{train_T20MTL}}
		\subfigure[STP]
		{\centering\scalebox{0.7}
			{\includegraphics{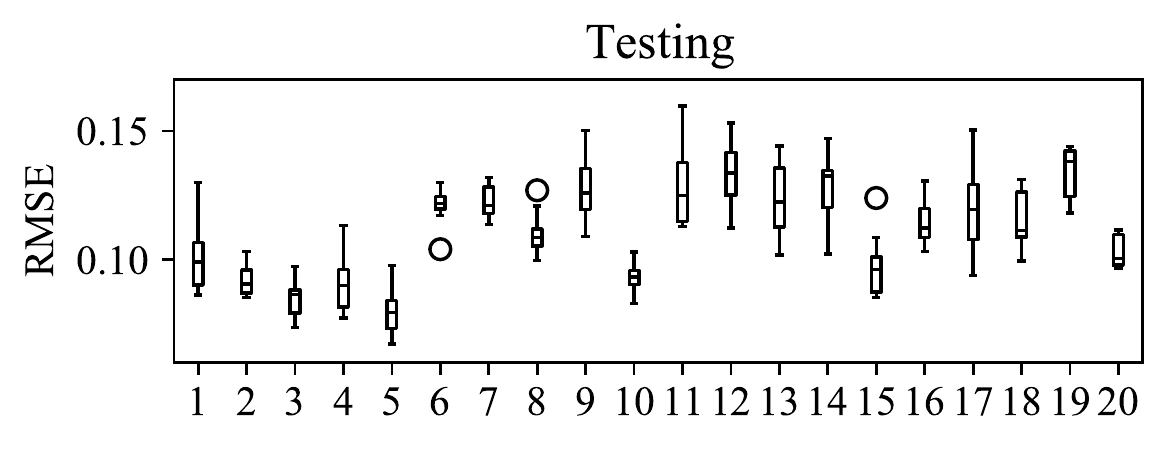}}\label{test_T20STL}}
		\subfigure[MTO-CT]
		{\centering\scalebox{0.7}
			{\includegraphics{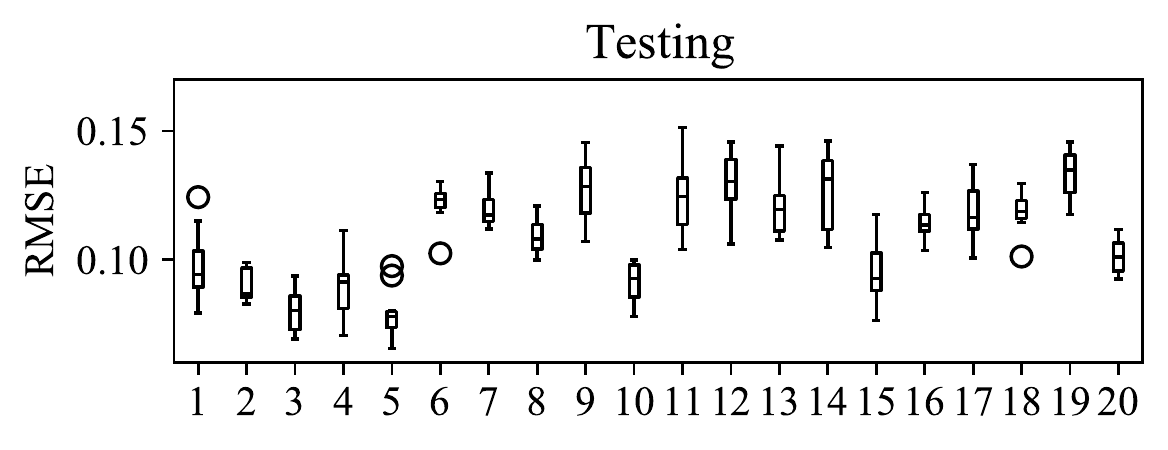}}\label{test_T20MTL}} 
		\caption{Box plots of training and testing RMSE on 20 tasks under STP and MTO-CT over ten independent runs.} \label{T20_distribution}
	\end{figure}
	
	Fig.~\ref{T5_distribution} illustrates the training and testing RMSE distribution of ten independent runs over each task for STP and MTO-CT. Fig.~\ref{train_T5STL} and Fig.~\ref{train_T5MTL} show that STP and MTO-CT have slight difference over the training performance from the view of distribution, but we still can see that the training RMSE on MTO-CT is better than STP. Even though MTO-CT has more outliers, fewer results deviate from the average value for each of the tasks. Similarly, comparing Fig.~\ref{test_T5STL} and Fig.~\ref{test_T5MTL}, the results on STP have more differences, resulting in higher average RMSE on each task, verified by the results in Table.~\ref{result_5task}. 
	
	\subsection{Results of Set B}
	
	Table.~\ref{result_20task} shows the average RMSE across ten independent runs over training and testing sets for STP and MTO-CT. The results are from 20 tasks that include 6-step, 12-step, 18-step, and 24-step ahead prediction over five states (the representations of the task numbers can be found in Table.~\ref{Task_number}). From Table.~\ref{result_20task}, we can see some tasks over MTO-CT have worse training performance than STP such as tasks 8 (12-step ahead prediction in SA), 9 (12-step ahead prediction in QLD), 14 (18-step ahead prediction in QLD), 17 (24-step ahead prediction in NSW), and 18 (24-step ahead prediction in SA). MTO-CT and STP have the same significant level on task 3 (6-step ahead prediction in SA) on the training set from the result of the statistical test, even though the average RMSE is slightly different. Among all 20 tasks, MTO-CT outperforms STP on 14 tasks on the training set. For the testing set, MTO-CT leads to better performance on 17 tasks than STP, where tasks 6 (12-step ahead prediction in VIC), 9, and 18 have worse performance (higher mean RMSE) with MTO-CT. The result of addressing 20 tasks simultaneously with inter-task knowledge transfer further demonstrates the superiority of MTO-CT, given that it outperforms STP on 14 and 17 tasks for training and testing sets, respectively.   
	
	Fig.~\ref{T20_distribution} illustrates the distribution of the training and testing RMSE on 20 tasks under STP and MTO-CT. For most of the tasks, Fig.~\ref{train_T20STL} and Fig.~\ref{train_T20MTL} show similar distribution over ten runs, except for task 18, which is significantly different and also has worse performance on MTO-CT. For the testing RMSE as presented in Fig.~\ref{test_T20STL} and Fig.~\ref{test_T20MTL}, task 18 on MTO-CT cannot compete with STP as well. However, for most of the rest, MTO-CT outperforms STP, further verified by Table.~\ref{result_20task}. Therefore, knowledge transfer among tasks in MTO-CT leads to better prediction performance for most tasks.

	\section{Conclusions and Future Work}\label{conclusion}
	
	We proposed an MTO-CT framework to solve multiple prediction tasks simultaneously, where an inter-task knowledge transfer module is designed to transfer and share knowledge among different tasks so that the overall performance of solving each task can be improved. MTO-CT employs an LSTM based model as the predictor and represents the knowledge as the connection weights and biases in LSTM. The inter-task knowledge transfer module is responsible for selecting the source tasks (w.r.t. a target task) from which the knowledge is extracted, extracting the knowledge, and reusing the extracted knowledge in the target task. The performance of MTO-CT is tested on two sets of tasks at different scales, i.e., five tasks and 20 tasks. The superiority of MTO-CT in terms of prediction accuracy is demonstrated in comparison to STP which solves each task in a standalone way without inter-task knowledge transfer. Our future work includes enriching the input by incorporating additional time series data like temperature, evaluating the performance of MTO-CT for co-training more LSTM variants or other types of prediction models~\cite{QIN2005773}, and applying MTO-CT to other applications that we worked on previously like graph matching~\cite{GONG2016158}, feature extraction~\cite{QIN2005613} service composition~\cite{Mistry2018131}.

\end{document}